\documentclass[letterpaper, 10 pt, conference]{ieeeconf}

\IEEEoverridecommandlockouts                              

\usepackage[T1]{fontenc}

\usepackage{amsfonts}	
\usepackage{amsmath}	
\usepackage{amssymb}    
\usepackage{siunitx}
\usepackage{pifont}   
\usepackage{dsfont}   
\usepackage{caption}
\usepackage{booktabs}
\usepackage{makecell}  
\usepackage[flushleft]{threeparttable}  
\usepackage{multirow}
\usepackage{rotating}

\usepackage{xspace}    
\usepackage[dvipsnames]{xcolor}    
\usepackage{colortbl}
\let\labelindent\relax
\usepackage[inline]{enumitem} 
\usepackage{textcomp}  
\usepackage{gensymb}   
\usepackage{graphicx} 
\usepackage[caption=false,font=footnotesize]{subfig}
\usepackage{microtype}
\usepackage{cite}
\usepackage{tikz}
\usepackage{array}
\usepackage{balance}

\makeatletter
\let\NAT@parse\undefined
\makeatother

\usepackage{url}

\usepackage[pdfencoding=auto, hidelinks]{hyperref} 
\usepackage[hang,flushmargin]{footmisc}

\usepackage[capitalize]{cleveref}
\crefname{section}{Sec.}{Secs.}
\Crefname{section}{Section}{Sections}
\Crefname{table}{Table}{Tables}
\crefname{table}{Tab.}{Tabs.}

\usepackage{balance}

\definecolor{Gray}{gray}{0.9}

\begin{document}

\title{\LARGE \bf
Leveraging Vision-Based Point Cloud Map Priors for Camera-Based 3D Object Detection and Online Vectorized HD Mapping
}

\author{
Markus Käppeler,
Rohit Mohan,
and Abhinav Valada
\thanks{Department of Computer Science, University of Freiburg, Germany.}%
\thanks{This work was funded by the European Union’s Horizon research innovation program grant No 101202228-HIDDEN and the German Research Foundation Emmy Noether Program grant number 468878300.}%
}

\maketitle


\begin{abstract}
    Camera-based 3D object detection and online vectorized HD mapping provide compact scene representations for autonomous driving, but both depend on accurate metric geometry and remain limited by depth ambiguity. Over long-term deployment, observations from repeated traversals can be accumulated into persistent point cloud priors that provide geometric context beyond the current observations. Existing explicit point cloud prior approaches, however, rely on LiDAR-based map construction and therefore require expensive 3D ranging sensors. We propose a framework that constructs a static point cloud prior map from previous camera traversals using Pi3X and augments each point with DINOv3 features. At runtime, a local prior patch is retrieved using global localization, encoded with a sparse voxel backbone, and fused in bird's-eye view (BEV) with lifted multi-view camera features. Task-specific sparse transformer heads then predict 3D objects and vectorized map elements from the fused representation. On Argoverse~2, the vision-based prior improves a strong baseline from 0.287 to 0.299 CDS and from 0.669 to 0.750 vectorized mapping mAP. Ablations show that semantic DINOv3 features are particularly important for vectorized mapping. These results demonstrate that vision-built geometric-semantic priors provide an effective form of long-term scene memory for camera-based perception, improving both tasks without LiDAR for prior-map construction or online inference.

\end{abstract}


\section{Introduction}

\label{sec:introduction}

Camera-based 3D object detection and online vectorized HD mapping are central to autonomous driving, as they provide the object- and map-level scene representations required for downstream prediction and planning~\cite{sun2024sparsedrive,cusumano2025robust,buchner2023learning}. LiDAR-only and camera--LiDAR fusion methods~\cite{yin2021center,lang2024point,yan2018second,liu2022bevfusion,mohan2024progressive,liao2025maptrv2,mohan2026up} achieve strong performance by exploiting accurate 3D geometry, but require expensive ranging sensors at inference. Camera-only approaches~\cite{kappeler2025bridging,lin2023sparse4dv3,wang2023exploring,li2024bevnext,sun2024sparsedrive,chen2024maptracker} are more scalable and retain rich semantic and high-resolution visual cues, yet remain fundamentally challenged by depth ambiguity~\cite{li2023bevdepth, mohan2024syn}. This is particularly limiting for vectorized mapping, where the model must infer static map structure despite occlusions caused by traffic participants and other scene elements~\cite{liao2025maptrv2,chen2024maptracker,luz2024amodal,sekkat2024amodalsynthdrive}.

During long-term deployment, autonomous vehicles repeatedly traverse the same environments including under open-set conditions~\cite{mohan2024panoptic,mohan2025open}, allowing observations across multiple agents and previous traversals to be accumulated into a persistent spatial prior with broad coverage. This accumulated prior map can then be retrieved for online perception to incorporate geometric prior knowledge as a complementary source beyond the current observations.
Previous work has exploited such priors for 3D occupancy prediction using neural rendering-based maps~\cite{yuan2024presight,luz2026latent} or global occupancy maps~\cite{yuan2025lmpocc}. PreSight~\cite{yuan2024presight} additionally applies its reconstructed prior to vectorized mapping, but requires computationally heavy per-scene NeRF optimization. For 3D object detection, Hindsight~\cite{you2022hindsight} uses prior LiDAR maps for LiDAR-based perception, while AsyncDepth~\cite{you2024better} and DualViewMapDet~\cite{kappeler2026leveraging} introduce LiDAR-map priors for camera-based 3D object detection. While these detection methods demonstrate the value of previous-traversal geometry for 3D detection, existing explicit point cloud priors for camera-based methods are still constructed from LiDAR traversals.

In this work, we investigate whether an explicit point cloud prior can instead be built from previous camera traversals without LiDAR-based mapping and benefit \emph{both} 3D object detection and vectorized mapping. We construct such a prior using the feed-forward Pi3X reconstruction model~\cite{wang2025pi} and augment each reconstructed point with dense DINOv3 features~\cite{simeoni2025dinov3}. The resulting geometric-semantic representation combines metric scene structure with image-derived semantic features in a common 3D map. The reconstructed geometry provides context that can reduce depth ambiguity for 3D object detection, while the accumulated prior retains information about static map elements that may be occluded in the current observations.

\begin{figure}
    \centering
    \includegraphics[width=0.9\linewidth]{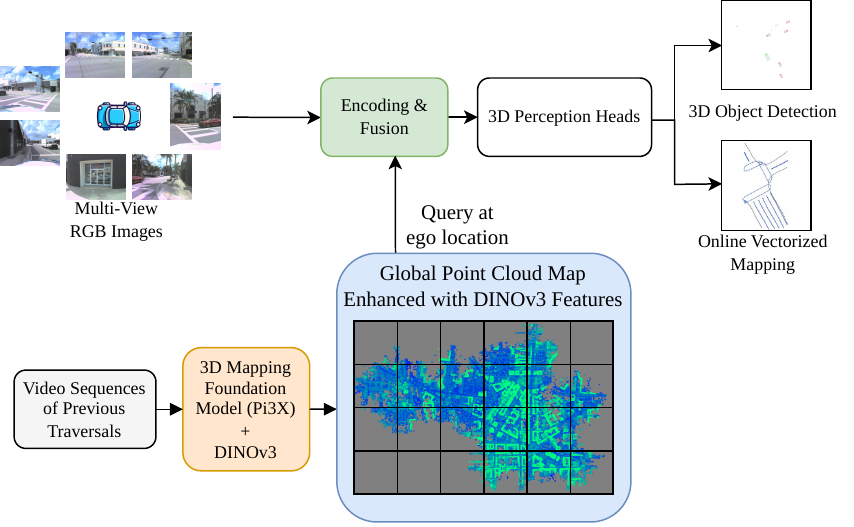}
    \caption{
    Camera-map perception with a vision-built prior. Given multi-view images and the current ego pose, we retrieve a local DINOv3-augmented point cloud reconstructed from earlier camera traversals. The prior map is encoded into bird's-eye view (BEV) and fused with lifted image features. Task-specific transformer heads then predict 3D bounding boxes and vectorized HD map elements from the fused BEV features.}
    \label{fig:overview-pi3xmapdet}
\end{figure}

We propose a camera-based framework that retrieves a local patch from a global vision-built prior map, encodes it with a sparse voxel backbone, and fuses it with camera BEV features obtained through LSS lifting~\cite{philion2020lift}, as shown in Fig.~\ref{fig:overview-pi3xmapdet}. The fused representation is decoded with SparseDrive-style transformer heads~\cite{sun2024sparsedrive} for 3D object detection and online vectorized HD mapping, while perspective-view (PV) image features remain available to both heads through PV deformable aggregation. At inference, our method requires only calibrated multi-view RGB cameras, global localization, and access to the offline prior map. No LiDAR is required for map construction or online perception. Experiments on Argoverse~2 show improvements for both tasks, with particularly strong gains in vectorized mapping when the prior is enriched with DINOv3 features. These results demonstrate the potential of persistent cross-traversal priors to improve perception during subsequent visits using information accumulated over long-term deployment.

\section{Technical Approach}
\label{sec:method}

\begin{figure*}
    \centering
    \includegraphics[width=0.75\linewidth]{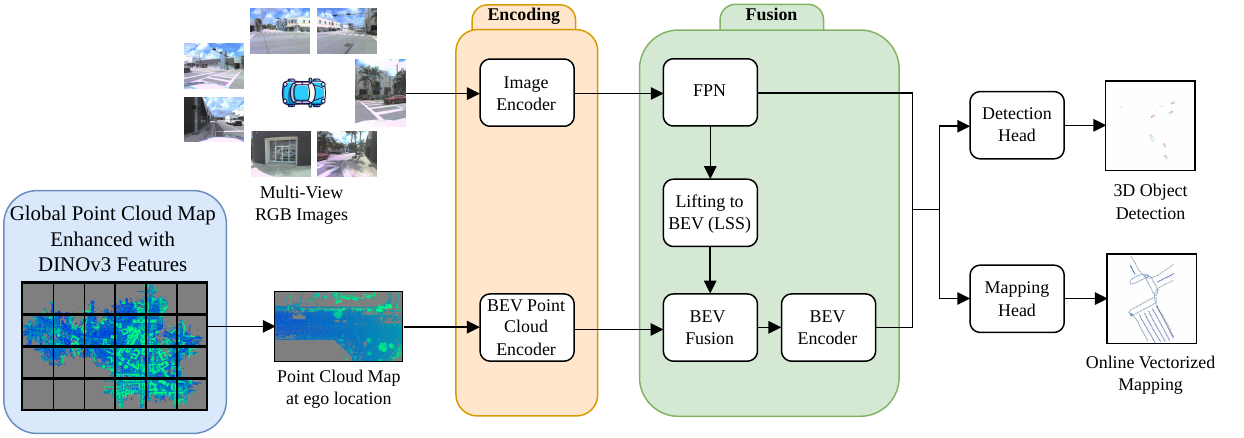}
    \caption{Overview of our approach. Multi-view images are encoded into perspective-view (PV) features and lifted to BEV via LSS, while a locally built vision-based point cloud prior is retrieved at the ego location and encoded by a BEV point cloud encoder. The camera and map BEV features are fused in a shared metric space. Sparse transformer heads for 3D detection and vectorized mapping then sequentially aggregate fused BEV features and multi-view PV image features.}
    \label{fig:overview}
\end{figure*}

We propose a camera-based framework for joint 3D object detection and vectorized mapping that exploits an offline static point cloud \emph{prior map} reconstructed from previous camera traversals (Fig.~\ref{fig:overview}). At inference, the model requires calibrated multi-view RGB images, a globally localized ego pose, and retrieval of a local prior-map patch. Unlike previous prior-map methods based on LiDAR~\cite{you2024better,kappeler2026leveraging}, neither map construction nor online perception requires LiDAR sensing to enable scalable deployment.
 
\subsection{Problem Setup}
For each time step $t$, the input consists of synchronized surround-view images $\{I_t^c\}_{c=1}^{N}$, their calibration parameters, and a globally localized ego pose $T^{G}_{E_t}\in SE(3)$. An offline global static map $\mathcal{M}^{G}$ is constructed from previous traversals. The model predicts oriented 3D boxes
\begin{equation}
\mathcal{B}_t = \{ b_i^t \}_{i=1}^{M_d},\qquad
b_i^t=(x,y,z,w,h,l,\mathrm{yaw},\mathbf{v}),
\end{equation}
and a set of vectorized map elements represented as polylines~\cite{sun2024sparsedrive}
\begin{equation}
\mathcal{L}_t=\{l_i^t\}_{i=1}^{M_m},\qquad
l_i^t=((x_{i,k},y_{i,k}))_{k=1}^{N_p},
\end{equation}
with $N_p=20$ points per road boundary, lane divider, or pedestrian crossing. Following SparseDrive~\cite{sun2024sparsedrive}, detection is performed within a \SI{50}{\meter} radius and mapping within a $\SI{30}{\meter}\times\SI{60}{\meter}$ ego-centered region.

\subsection{Vision-Based Prior Map Construction and Retrieval}
\label{ssec:method-map}

\paragraph*{Pi3X reconstruction.}
We use Pi3X, the enhanced $\pi^3$ feed-forward reconstruction model~\cite{wang2025pi}, to build the prior map from camera data. Given a set of input images, Pi3X jointly predicts camera poses and intrinsics, local 3D point maps, and point confidences. Each local point map assigns every image pixel a 3D point in the coordinate system of the corresponding camera. We process each sequence in chunks of 12 multi-view frames and transform the local point maps from all cameras and time steps into a common global frame using the dataset-provided camera poses. The transformed point maps are then concatenated to accumulate all point clouds for a chunk. 

Because the reconstruction scale can vary across chunks, we recover metric scale using the dataset trajectory. Specifically, we estimate a scale factor from the displacement between the first and last Pi3X-predicted poses and the corresponding dataset-provided global poses,
\begin{equation}
\alpha=
\frac{\lVert\mathbf{t}^{G}_{K}-\mathbf{t}^{G}_{1}\rVert_2}
{\lVert\hat{\mathbf{t}}_{K}-\hat{\mathbf{t}}_{1}\rVert_2},
\end{equation}
and apply $\alpha$ to the predicted local depth before transforming the reconstructed points into the global coordinate frame using the dataset-provided camera poses. Hence, the Pi3X-predicted poses are used only to recover metric scale, while the dataset poses establish the relative alignment between frames and the global reference frame.

We retain points with a Pi3X confidence score of at least $0.5$, remove points closer than \SI{0.6}{\meter} to their corresponding camera or farther than \SI{60}{\meter}, and mask out dynamic objects using projected 3D boxes. The remaining static points are average-pooled in \SI{0.2}{\meter} voxels and stored in a tiled global map as in DualViewMapDet~\cite{kappeler2026leveraging}.

\paragraph*{DINOv3 semantic features.}
Geometry alone is noisy and less informative than a semantic map prior. We therefore extract dense fine-grained DINOv3 ViT-S features~\cite{simeoni2025dinov3} using the multi-scale procedure of~\cite{vodisch2024good}. Pretrained vision models have also been adapted to LiDAR
semantic segmentation~\cite{hindel2025label} and hyperspectral
semantic segmentation~\cite{hurtado2026hyperspectral}. Since the Pi3X local point map and DINOv3 feature map are image aligned, each reconstructed point is directly assigned the feature at its corresponding pixel. We fit a PCA projection on a large feature subset and compress the original 384-dimensional descriptors to 64 dimensions before storage. Each map point thus carries its 3D position, Pi3X confidence, RGB value, and compressed DINOv3 descriptor.

\paragraph*{Online retrieval}
Given the current ego translation $\mathbf{t}^{G}_{E_t}$ and retrieval range $R$, we query a local global-map patch
\begin{equation}
\mathcal{P}_t^{G}=\left\{\mathbf{p}^{G}\in\mathcal{M}^{G}\ \middle|\ 
\lVert(p_x^{G},p_y^{G})-(t_x^{G},t_y^{G})\rVert_{\infty}\le R\right\}.
\end{equation}
To avoid leakage, points from the current traversal and temporally adjacent traversals are excluded. The retrieved points from multiple traversals are merged and transformed to the current ego frame:
\begin{equation}
\mathcal{P}_t^{E}=\left\{(T^{G}_{E_t})^{-1}\mathbf{p}^{G}\ \middle|\ \mathbf{p}^{G}\in\mathcal{P}_t^{G}\right\}.
\end{equation}

\subsection{Camera--Map Encoding and Fusion}
We extract multi-scale PV image features
\begin{equation}
\mathbf{F}^{c,s}_{\mathrm{img}}=\phi^{s}_{\mathrm{img}}(I_t^c),\qquad s\in\mathcal{S},
\end{equation}
with VoVNet-99~\cite{lee2019energy}. An LSS-style lifting module~\cite{philion2020lift,li2024bevnext,mohan2026forecastocc} predicts per-pixel depth distributions and pools the corresponding frustum features from all cameras into $\mathbf{F}^{\mathrm{cam}}_{\mathrm{BEV}}$.

For the prior map, each ego-frame point $j$ is represented by
\begin{equation}
\mathbf{x}_j=\left[\mathbf{p}^{E}_j\ \Vert\ q_j\ \Vert\ \mathbf{r}_j\ \Vert\ \mathbf{d}_j\right],
\end{equation}
where $q_j$ is Pi3X confidence, $\mathbf{r}_j$ denotes RGB, and $\mathbf{d}_j\in\mathbb{R}^{64}$ is the compressed DINOv3 descriptor. We voxelize the points at \SI{0.2}{\meter} resolution, average their attributes within each voxel, and encode them using a sparse 3D backbone~\cite{yan2018second}:
\begin{equation}
\mathbf{F}^{\mathrm{map}}_{\mathrm{BEV}}
=\phi_{\mathrm{map}}\!\left(\mathrm{Voxelize}(\mathcal{P}_t^{E})\right).
\end{equation}
The map and camera BEV features have the same spatial resolution and are fused by channel-wise concatenation followed by a $3\times3$ Conv-BN-ReLU block~\cite{liu2022bevfusion,kappeler2026leveraging}:
\begin{equation}
\mathbf{F}_{\mathrm{BEV}}=
\psi_{\mathrm{BEV}}\!\left([\mathbf{F}^{\mathrm{cam}}_{\mathrm{BEV}}\Vert\mathbf{F}^{\mathrm{map}}_{\mathrm{BEV}}]\right),
\end{equation}
followed by lightweight ResNet blocks and an FPN to refine the fused metric representation.

\subsection{Sparse Detection and Mapping Heads}
We follow the symmetric sparse perception design of SparseDrive~\cite{sun2024sparsedrive}, augmented with the sequential BEV--PV deformable aggregation~\cite{kappeler2026leveraging, kappeler2025bridging}. The detection branch maintains 3D box anchors $A^{d}\in\mathbb{R}^{M_d\times11}$ and instance features $F^{d}\in\mathbb{R}^{M_d\times C}$, while the mapping branch represents anchors as polylines $L^{m}\in\mathbb{R}^{M_m\times N_p\times2}$ with corresponding instance features $F^{m}\in\mathbb{R}^{M_m\times C}$~\cite{sun2024sparsedrive}. In each decoder layer, both branches first aggregate the fused BEV representation and then the multi-scale PV image features:
\begin{equation}
\begin{aligned}
\mathbf{f}'_i &= \mathbf{f}_i + \mathrm{BEV\text{-}DeformAgg}(\mathbf{f}_i,\mathbf{a}_i,\mathbf{F}_{\mathrm{BEV}}),\\
\mathbf{f}''_i &= \mathbf{f}'_i + \mathrm{PV\text{-}DeformAgg}(\mathbf{f}'_i,\mathbf{a}_i,\{\mathbf{F}^{c,s}_{\mathrm{img}}\}).
\end{aligned}
\end{equation}
For detection, deformable keypoints are sampled from the 3D anchor box. For mapping, the polyline points serve as the geometric sampling locations. This exposes both tasks to the metric camera--map context in BEV while retaining fine-grained image information in PV.

\subsection{Training Objective}
We train the complete network end-to-end with the standard SparseDrive detection and mapping objectives~\cite{sun2024sparsedrive} and LiDAR depth supervision for the LSS module~\cite{li2023bevdepth,li2024bevnext}:
\begin{equation}
\mathcal{L}_{\mathrm{total}}=
\lambda_{\mathrm{det}}\mathcal{L}_{\mathrm{det}}+
\lambda_{\mathrm{map}}\mathcal{L}_{\mathrm{map}}+
\lambda_{\mathrm{depth}}\mathcal{L}_{\mathrm{depth}}.
\end{equation}
The prior-map encoder is optimized only through the downstream task losses. We additionally apply image--map grid masking during training as in DualViewMapDet~\cite{kappeler2026leveraging} to reduce overreliance on locally available prior maps.

\section{Experimental Evaluation}
\label{sec:result}

We evaluate our method on Argoverse~2 to quantify how vision-built prior accumulated across previous traversals affects camera-based 3D object detection and online vectorized HD mapping. We compare against strong camera-based baselines under standard evaluation protocols and ablate the contributions of prior geometry and DINOv3 semantics.

\subsection{Implementation and Training Details}
The image backbone produces multi-scale features at strides $\{4,8,16,32\}$, and the camera and map BEV branches use a shared $128\times128$ spatial grid. We lift stride-$8$ PV features to $480\times704$ resolution and stride-$16$ features for the higher-resolution setting. Following our baselines, VoVNet-99~\cite{lee2019energy} is initialized from FCOS3D pretraining on nuScenes. We train for 80 epochs with AdamW, a base learning rate of $2\times10^{-4}$, cosine decay, and a 500-iteration warm-up. The batch size is 24, and we use sequential iteration training. Image and BEV/3D augmentations follow~\cite{kappeler2025bridging}. The LSS depth estimator is supervised with LiDAR depth during training, while inference remains camera-based. We additionally employ image--map grid masking~\cite{kappeler2026leveraging}.

\subsection{Dataset and Metrics}
Argoverse~2~\cite{wilson2023argoverse} provides synchronized surround-view cameras and repeated coverage of urban areas, enabling prior-map construction from other traversals. Training maps are built exclusively from the training split. For validation, maps may contain training and validation traversals, but never the current traversal. We use the globally corrected Argoverse~2 poses from DualViewMapDet~\cite{kappeler2026leveraging}. For 3D detection, we report the official Composite Detection Score (CDS), mAP, and true-positive errors mATE, mASE, and mAOE over the 26 classes. Vectorized mapping is evaluated with mAP following the protocol used by StreamMapNet~\cite{yuan2024streammapnet} and MapTracker~\cite{chen2024maptracker}.

\subsection{3D Object Detection and Vectorized Mapping}

\begin{table*}
\footnotesize
\centering
\caption{3D object detection and online vectorized HD mapping results on Argoverse~2 val set.}
\label{tab:results-argoverse2}
\setlength\tabcolsep{2.5pt}
\begin{threeparttable}
    \begin{tabular}{l | c c c | c c | c c c c c | c }
        \toprule
        & & & & & & \multicolumn{5}{c}{\textbf{3D Object Detection}} & \textbf{\makecell{Vectorized \\HD Mapping}} \\
        \textbf{Method} & \textbf{3D Det.} & \textbf{Vec. Map.} & \textbf{Map Prior} & \textbf{Backbone} & \textbf{Image Size} & \textbf{CDS$\uparrow$} & \textbf{mAP$\uparrow$} & \textbf{mATE$\downarrow$} & \textbf{mASE$\downarrow$} & \textbf{mAOE$\downarrow$} & \textbf{mAP$\uparrow$} \\
        \midrule
        Far3D\textsuperscript{*}~\cite{jiang2024far3d} & \checkmark & & & VoVNet-99 & $640 \times 960$ & 0.281 & 0.367 & 0.730 & 0.300 & 0.531 & -- \\
        Sparse4Dv3\textsuperscript{*}\textsuperscript{\textdaggerdbl}~\cite{lin2023sparse4dv3} & \checkmark & & & VoVNet-99 & $640 \times 960$ & 0.294 & 0.381 & 0.706 & 0.314 & 0.666 & -- \\
        Sparse4Dv3\textsuperscript{\S}\textsuperscript{\textdaggerdbl}~\cite{lin2023sparse4dv3} & \checkmark & & & VoVNet-99 & $640 \times 960$ & 0.288 & 0.376 & 0.694 & 0.298 & 0.547 & -- \\
        DualViewMapDet \textsuperscript{\S}~\cite{kappeler2026leveraging} & \checkmark & & LiDAR & VoVNet-99 & $640 \times 960$ & 0.311 & 0.397 & 0.622 & 0.279 & 0.527 & -- \\
        \rowcolor{Gray} Ours\textsuperscript{\S} & \checkmark & \checkmark & Vision & VoVNet-99 & $640 \times 960$ & 0.314 & 0.413 & 0.662 & 0.290 & 0.599 & 0.753 \\
        \midrule
        StreamMapNet\textsuperscript{\S}~\cite{yuan2024streammapnet} & & \checkmark & & ResNet50 & $480 \times 800$ & -- & -- & -- & -- & -- & 0.640 \\
        SQD-MapNet\textsuperscript{\S}~\cite{wang2024stream} & & \checkmark & & ResNet50 & $480 \times 800$ & -- & -- & -- & -- & -- & 0.633 \\
        MapTracker\textsuperscript{\textdagger}~\cite{chen2024maptracker} & & \checkmark & & ResNet50 & $480 \times 800$ & -- & -- & -- & -- & -- & 0.714 \\
        \midrule
        \makecell{SparseDrive\textsuperscript{\S}\textsuperscript{\textdaggerdbl}~\cite{sun2024sparsedrive} (+ BEV net. \\ \& 3D/BEV data aug.\cite{kappeler2025bridging})} & \checkmark & \checkmark & & VoVNet-99 & $480 \times 704$ & 0.287 & 0.381 & 0.709 & 0.287 & 0.668 & 0.669  \\
        \rowcolor{Gray} Ours\textsuperscript{\S} & \checkmark & \checkmark & Vision & VoVNet-99 & $480 \times 704$ & 0.299 & 0.396 & 0.696 & 0.292 & 0.641 & 0.750 \\ 
        \bottomrule
    \end{tabular}
    \footnotesize
    We evaluate object detection within a range of 50 meters, and vectorized mapping uses a $\SI{30}{\meter}\times\SI{60}{\meter}$ region.
    \textsuperscript{*}:~Training uses the $\SI{10}{\hertz}$ Argoverse~2 sequences by splitting each sequence into five offset subsequences, yielding \(\approx 5\times\) more (but redundant) training samples than strict $\SI{2}{\hertz}$ subsampling.
    \textsuperscript{\S}:~Training with strict $\SI{2}{\hertz}$ subsampling.
    \textsuperscript{\textdagger}:~Training with strict $\SI{2.5}{\hertz}$ subsampling.
    \textsuperscript{\textdaggerdbl}:~Baselines trained by us with the code provided by the authors.
\end{threeparttable}
\end{table*}

\begin{table}[t]
\footnotesize
\centering
\caption{Ablation study of the prior map and DINOv3 features.}
\label{tab:results-ablation}
\setlength{\tabcolsep}{1.1pt}
\renewcommand{\arraystretch}{0.95}

\begin{tabular}{ccc | cc | c}
    \toprule
    &&&
    \multicolumn{2}{c|}{\makecell{\textbf{3D Object}\\\textbf{Detection}}} &
    \makecell{\textbf{Vectorized}\\\textbf{Mapping}} \\
    \cmidrule(lr){4-5}\cmidrule(lr){6-6}
    \textbf{Prior Map} &
    \textbf{DINOv3 Feat.} &
    \makecell{\textbf{All Prior}\\\textbf{Maps}} &
    \textbf{CDS$\uparrow$} &
    \textbf{mAP$\uparrow$} &
    \textbf{mAP$\uparrow$} \\
    \midrule
    &&& 0.287 & 0.381 & 0.669 \\
    \checkmark &&& 0.284 & 0.377 & 0.683 \\
    \checkmark & \checkmark && 0.291 & 0.388 & 0.756 \\
    \rowcolor{Gray}
    \checkmark & \checkmark & \checkmark & 0.299 & 0.396 & 0.750 \\
    \bottomrule
\end{tabular}

\vspace{0.7mm}
\parbox{\columnwidth}{\scriptsize
All variants use VoVNet-99 pretrained with FCOS3D at
$480\times704$. The baseline is SparseDrive~\cite{sun2024sparsedrive}
augmented with a BEV network, deformable BEV aggregation, and
3D/BEV data augmentations~\cite{kappeler2025bridging}.}
\end{table}

\begin{table}
\footnotesize
\centering
\caption{Results on subsets with higher prior-map coverage.}
\label{tab:results-argoverse2-map-coverage-subsets-net-baseline}
\setlength\tabcolsep{2.0pt}
\begin{threeparttable}
    \begin{tabular}{l | c c | c | c c | c}
        \toprule
        \multirow{2}{*}{\textbf{Subset}} &
        \multicolumn{3}{c|}{\textbf{Ours (map prior)}} &
        \multicolumn{3}{c}{\textbf{Baseline (no map)}} \\
        \cmidrule(lr){2-4}\cmidrule(lr){5-7}
        & \textbf{CDS$\uparrow$} & \textbf{mAP$\uparrow$} & \textbf{mAP$\uparrow$}
        & \textbf{CDS$\uparrow$} & \textbf{mAP$\uparrow$} & \textbf{mAP$\uparrow$} \\
        \midrule
        All & 0.299 {\color{ForestGreen}(+$1.2$\,pp)} & 0.396 & 0.750 & 0.287 & 0.381 & 0.669 \\
        Overlap $k{=}1$ & 0.331 {\color{ForestGreen}(+$1.9$\,pp)} & 0.436 & 0.505 & 0.312 & 0.409 & 0.440 \\
        Overlap $k{=}2$ & 0.315 {\color{ForestGreen}(+$1.8$\,pp)} & 0.414 & 0.378 & 0.296 & 0.391 & 0.328 \\
        \bottomrule
    \end{tabular}
    \footnotesize
    Overlap $k$: a sequence is included iff every $50\,\mathrm{m}\times50\,\mathrm{m}$ tile it covers is also covered by at least $k$ \emph{other} sequences. The map-free baseline is evaluated on the same subsets to isolate the effect of the prior. Gains (green) are in percentage points (pp) for CDS.
\end{threeparttable}
\end{table}

\cref{tab:results-argoverse2} compares our method with detection-only, mapping-only, and joint perception baselines. Under the directly matched $480\times704$ SparseDrive setting, the vision-based prior improves detection from 0.287 to 0.299 CDS and from 0.381 to 0.396 mAP, while reducing mATE from 0.709 to 0.696 and mAOE from 0.668 to 0.641. The gain is substantially larger for vectorized mapping, where mAP increases from 0.669 to 0.750. Thus, although the reconstructed map is less geometrically accurate than a LiDAR prior, its accumulated static structure and semantic features provide useful context for both object- and map-level perception.

\subsection{Ablation Studies}

{\parskip=2pt
\noindent\textit{Prior map and DINOv3 features:}
\cref{tab:results-ablation} isolates the contribution of the prior representation. Geometry alone slightly improves mapping mAP from 0.669 to 0.683, but does not improve detection, indicating that noisy camera-reconstructed geometry is not sufficient by itself. Augmenting the points with DINOv3 features raises detection to 0.291 CDS/0.388 mAP and mapping to 0.756 mAP. The particularly large mapping gain indicates that semantic foundation-model features provide strong cues for static structures such as lane dividers, road boundaries, and pedestrian crossings. Using prior maps from all prior traversals instead of only the two traversals with the largest point clouds per tile further improves detection to 0.299 CDS and 0.396 mAP, while mapping remains comparable at 0.750 mAP.
}

{\parskip=2pt
\noindent\textit{Map coverage:}
Not every validation sequence has equally strong prior coverage. We therefore evaluate subsets with increasing overlap and re-evaluate the map-free baseline on exactly the same samples in~\cref{tab:results-argoverse2-map-coverage-subsets-net-baseline}. The prior remains beneficial on all subsets, and the CDS improvement is larger on the coverage-restricted subsets ($+1.8$--$1.9$ pp) than over the full validation set ($+1.2$ pp), confirming that reliable previous-traversal coverage is important for exploiting the map prior.
}

\section{Conclusion}
We presented a framework for joint camera-based 3D object detection and online vectorized HD mapping that constructs a persistent geometric-semantic point-cloud prior from previous camera traversals. Pi3X geometry is enriched with DINOv3 features, encoded in BEV, and fused with camera features before sparse task-specific decoding. Experiments on Argoverse~2 show consistent gains over a strong SparseDrive-style baseline, with the largest improvements for vectorized mapping when semantic DINOv3 features are included. These results show that observations accumulated over long-term deployment can be retained as an explicit cross-traversal spatial prior and improve perception during subsequent visits, without requiring LiDAR for prior-map construction or online inference. In future work, we aim to improve camera--map fusion through better spatial alignment in BEV and stronger alignment of the two modalities in feature space. We will further investigate map change detection to increase robustness to outdated prior maps and to automatically identify when map updates are required.

\balance
\footnotesize
\bibliographystyle{IEEEtran}
\bibliography{references.bib}


\end{document}